\documentclass[acmtog,nonacm]{acmart}
\makeatletter
\def\@ACM@checkaffil{
    \if@ACM@instpresent\else
    \ClassWarningNoLine{\@classname}{No institution present for an affiliation}%
    \fi
    \if@ACM@citypresent\else
    \ClassWarningNoLine{\@classname}{No city present for an affiliation}%
    \fi
    \if@ACM@countrypresent\else
        \ClassWarningNoLine{\@classname}{No country present for an affiliation}%
    \fi
}
\makeatother

\usepackage{booktabs} 

\usepackage{microtype}
\usepackage{graphicx}
\usepackage{subfigure}
\usepackage{caption}
\usepackage{bbm}
\usepackage{url}
\usepackage{multirow,makecell, rotating}
\usepackage{tikz}
\usepackage{graphbox}
\usepackage{comment}
\usepackage{epsfig}
\usepackage{bm}
\usepackage{soul}
\usepackage{color}
\usepackage{hyperref}

\usepackage{amsmath}
\usepackage{mathtools}
\usepackage{amsthm}

\usepackage[ruled]{algorithm2e} 

\SetAlFnt{\small}
\SetAlCapFnt{\small}
\SetAlCapNameFnt{\small}
\SetAlCapHSkip{0pt}

\usepackage[capitalize]{cleveref}
\crefname{section}{Sec.}{Secs.}
\Crefname{section}{Section}{Sections}
\Crefname{table}{Table}{Tables}
\crefname{table}{Tab.}{Tabs.}

\begin{document}
\title{Face0: Instantaneously Conditioning a Text-to-Image Model on a Face}

\author{Dani Valevski}
\orcid{1234-5678-9012-3456}
\affiliation{%
 \institution{Google Research}
}
\email{daniv@google.com}
\authornote{Equal contribution.}

\author{Danny Wasserman}
\orcid{1234-5678-9012-3456}
\affiliation{%
 \institution{Google Research}
 }
\email{dwasserman@google.com}
\authornotemark[1]

\author{Yossi Matias}
\orcid{1234-5678-9012-3456}
\affiliation{%
 \institution{Google Research}
 }
\email{yossi@google.com}

\author{Yaniv Leviathan}
\orcid{1234-5678-9012-3456}
\affiliation{%
 \institution{Google Research}
 }
\email{leviathan@google.com}
\authornotemark[1]

\begin{abstract}
We present Face0, a novel way to instantaneously condition a text-to-image generation model on a face, in sample time, without any optimization procedures such as fine-tuning or inversions. We augment a dataset of annotated images with embeddings of the included faces and train an image generation model, on the augmented dataset. Once trained, our system is practically identical at inference time to the underlying base model, and is therefore able to generate images, given a user-supplied face image and a prompt, in just a couple of seconds. Our method achieves pleasing results, is remarkably simple, extremely fast, and equips the underlying model with new capabilities, like controlling the generated images both via text or via direct manipulation of the input face embeddings. In addition, when using a fixed random vector instead of a face embedding from a user supplied image, our method essentially solves the problem of consistent character generation across images. Finally, while requiring further research, we hope that our method, which decouples the model's textual biases from its biases on faces, might be a step towards some mitigation of biases in future text-to-image models.
\end{abstract}

\begin{teaserfigure}
\centering
\includegraphics[width=0.80\textwidth]{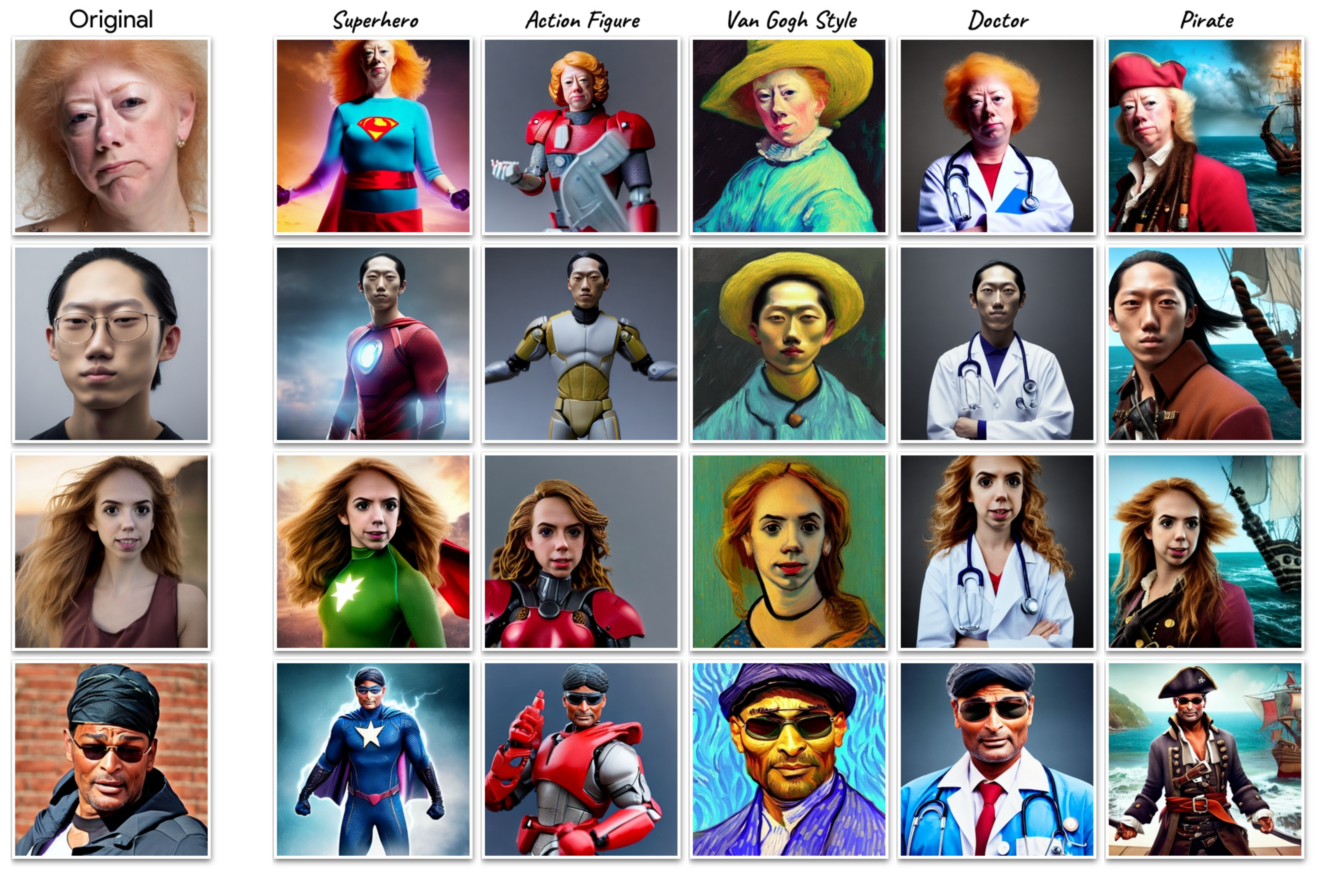}
    \vspace{-0.2cm}
    \caption{
   Example generations from Face0. It only takes a couple of seconds to generate an image given a single face image (left) and a textual prompt (top).
    }
    \label{fig:fig1}
\end{teaserfigure}

\maketitle

\begin{figure*}[ht]
\centering
\includegraphics[width=.8\linewidth]{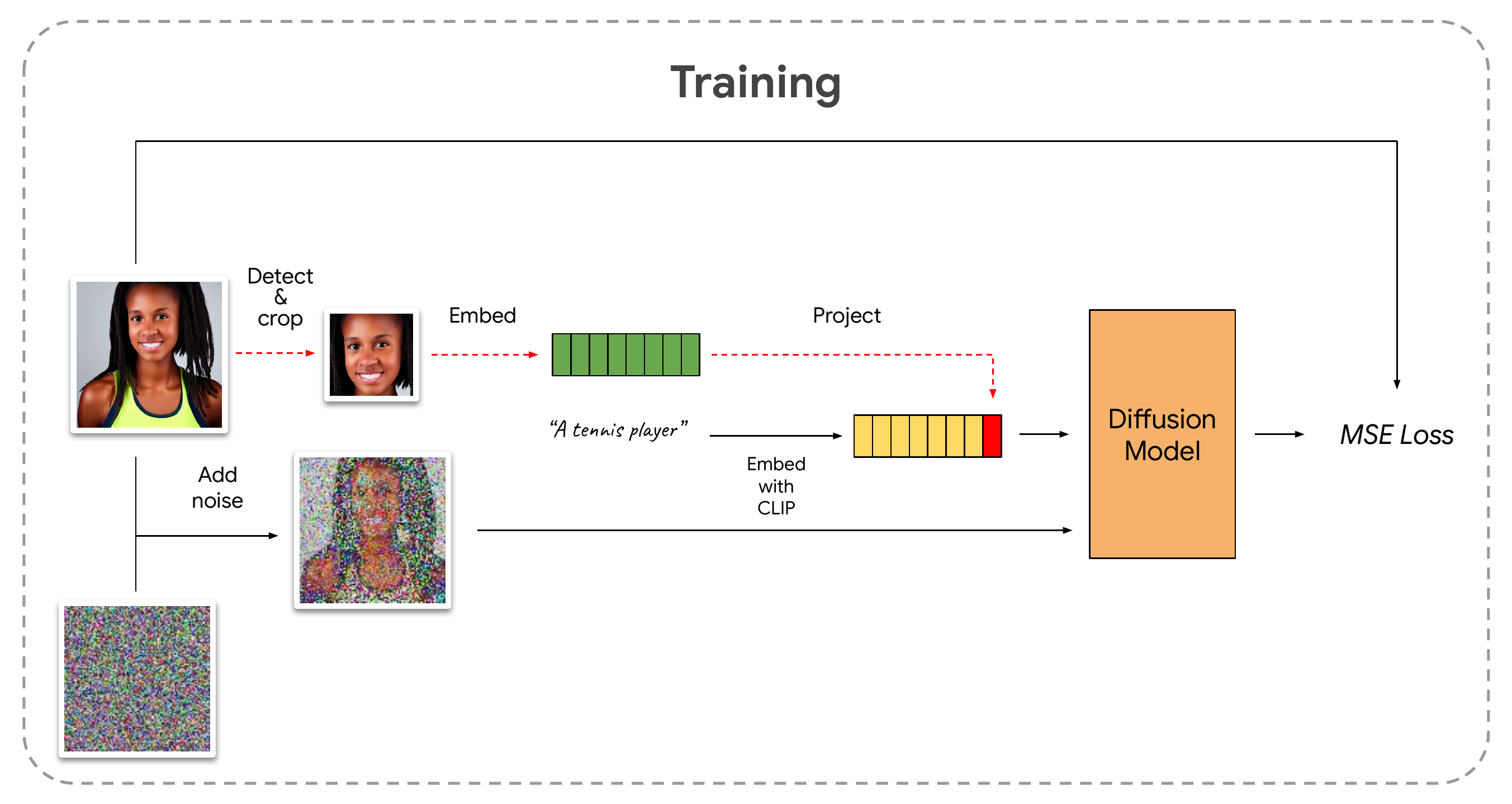}
\caption{The training scheme for Face0 (see \cref{sec:method}). Everything except the dashed red arrows is part of the standard diffusion model training procedure. For simplicity, we omit the details of converting from pixel space to latent space.}
\label{fig:system-diagram}
\end{figure*}

\section{Introduction}
The field of text-to-image synthesis has recently experienced rapid development, largely thanks to advances in diffusion models. By conditioning on free-form text, images of unprecedented quality and diversity can now be generated. However, generating an image depicting a person from a user-supplied image is still a challenging task. To overcome this gap, existing methods rely on solving an optimization problem during inference time, e.g. fine-tuning the model \cite{Dreambooth} or reversing the image into the textual embedding space \cite{TextualInversion}. While these methods produce good results they are costly in time or memory.

In this work we develop a novel method for instantaneously conditioning an image generation model, we use Stable Diffusion, on a face. At inference time our method is practically identical to standard inference from the base diffusion model, and enables instantly generating images in the likeness of a person from a single photo. For example, the images in \cref{fig:fig1} were generated in just a couple of seconds, given the original images on the left and the textual prompts at the top.

Face0 generates pleasing results (\cref{fig:fig1}), while being especially simple and efficient. In addition, it has several other advantages: (1) it enables easy and natural control of the generated faces, e.g. changing hair styles or orientations, both through textual prompts as well as more gradually through direct manipulation of the face embedding vectors (\cref{fig:prompt_control,fig:lerp,fig:text_and_face_lerp}), (2) by using \emph{fixed} randomized face embedding vectors, instead of a user-supplied face image, it trivially solves the problem of generating \emph{consistent} characters across generated images, and (3) since it encourages the model to decode the facial features from the face embedding, instead of from the textual prompt, it decouples some of the model's textual biases from its biases for facial features and, while more research is needed, we are hopeful that this is a step towards mitigating some of the model's inherent biases for facial features (\cref{fig:bias}).

Our core idea is to leverage a \emph{face embedding} model.
Specifically, we take our dataset of training images and augment those that contain a face with the embeddings of the face.
We use a simple module (a small MLP) to project the embeddings to Stable Diffusion's context's space, and then jointly train the base diffusion model and this projection module to generate images conditioned on the face embeddings (see \cref{fig:system-diagram}). In sampling time, we calculate the face embeddings from the user-supplied image, add it to Stable Diffusion's context, and sample an image in practically the standard way (we use a slightly modified classifier free guidance).

\section{Method}
\label{sec:method}

The main idea is to train the underlying diffusion model to be conditioned on both text and the output of an efficient face embedding mechanism.

\subsection{Architecture and Training}
For our face embedding module we use part of an Inception Resnet V1 model (i.e. we drop the last layers), trained on vggface2 \cite{vggface2}. The model, with the dropped layers, is not suitable for accurate identification, but it is able to preserve enough of the visual details needed for a high quality generation.
This embedding module mostly fixes the face pose and expression.
We augment a dataset of annotated images transforming each (image, caption) pair into a (image, caption, face-embedding) triplet. We then train a 4 layer MLP to convert from the face embedding space into the CLIP embedding space, and jointly fine-tune the underlying model to receive both the CLIP embedding and the projected face embedding as conditions. Specifically, we fine-tune the parameters $\theta$ of our model $M_{\theta}$ to optimize the standard diffusion model MSE loss objective, with an additional conditioning on the embedding of the face:

$$L(\theta) = E_{t, x_0, d, f, \epsilon}[w_t||M_{\theta}(\alpha_{t}x_0 + \sigma_{t}\epsilon, t, d, f) - \epsilon)^2||]$$

Where $M_{\theta}$ is the full model, including the projection MLP, parameterized by $\theta$, $t \sim U(0, 1)$, $\epsilon \sim N(0, 1)$, $\alpha_t$, $\sigma_t$ and $w_t$ are the diffusion noise parameters (see \cite{DDPM}), $x_0$ is an image sampled from the dataset, $d$ is its associated text condition, and finally $f$ is our newly introduced face embedding condition (see \cref{fig:system-diagram}).

\subsection{Sampling}

At sample time, given the user-supplied image, we run the same face extraction logic we used in training time, calculate the projected face embedding, and use it to override the last three tokens.

We then use a slight variation of classifier free guidance (CFG) \cite{ClassifierFreeGuidance}.
Similarly to standard CFG, we calculate the following linear combination for the unconditioned result (with a negative weight) and the conditioned result (with a positive weight).
Specifically, we evaluate the following standard CFG formula:

$$\epsilon_{t} = w \cdot \hat{\epsilon_t}(z_t, d, f) + (1 - w) \cdot \epsilon_t(z_t)$$

Where $d$ is the textual prompt and $f$ is the face embedding. In our experiments we use a classifier free guidance weight of $w = 7.5$.
Unlike standard classifier free guidance, to allow more refined control of the result, we use a weighted mean of three separate conditioned vectors. The conditioned vectors are those conditioned on the textual prompt $d$ alone, the face embedding $f$ alone, and their combination. We choose a parametrization where $c$ represents the relative weight of the combined vector, and $a$ represents the relative weight of the vector conditioned on the face-only from the remainder. Overall we have:

$$\hat{\epsilon_t}(z_t, d, f) = c \cdot \epsilon_t(z_t, d, f) + (1 - c) \cdot (a \cdot \epsilon_t(z_t, f) + (1 - a) \cdot \epsilon_t(z_t, d))$$

In practice, at least one of these three weight terms is always 0. See \cref{sec:sampling} and \cref{fig:cfg_weights_variance}.

Finally, we note that our method tends to maintain extremely high consistency to the face across generations (see \cref{fig:consistency}). If this is undesirable, adding a small amount of noise to the input embedding can increase variety.

\subsection{Details}
We train our model on the LAION dataset \cite{LAION}, filtered by an aesthetics threshold of 5.5, and we further filter it to only include images that include a face larger than 20 pixels. We note that this filtering operation might amplify biases in the dataset \cite{filtering_amplifies_bias}. This results in $\sim10M$ (image, caption) pairs. We use MTCNN \cite{MTCNN} for face detection. If an image has multiple faces we only take the largest one (images with multiple faces are useful for conditioning the model on multiple character, we leave this for future research).
To generate the face embeddings we crop the image based on the output of MTCNN and resize to a 160x160px square, preserving aspect ratio. We expand the rectangle returned by MTCNN to include some more details, such as the hairstyle. For the expansion, we manually picked the values of 10 pixels for the left and right margins, 33 pixels for the top margin and 15 pixels for the bottom margin. We have not tuned these choices further and used these numbers throughout all of our experiments. 
We then run our face embedding module, which outputs an embedding vector. To project the vector into the CLIP embedding space, we use a simple 4-layer feed-forward network with dimensions of 768 and ReLU non-linearities after the hidden layers. This results in $\sim10M$ additional total parameters.

The output of the projection is three 768-dimensional vectors which we use to override the last three tokens (tokens 75-77) in Stable Diffusion's CLIP embedding vector.
To support classifier free guidance, we zero out the projected embedding with a probability of 10\%. Note that we do not do the same for the text embedding, which might have improved the quality of the generations, and could be an interesting direction for further research.
We then jointly train the U-Net, starting from the Stable Diffusion 1.4 checkpoint, and the projection network, which is initialized randomly. We train on 64 TPU-v4s for 500K steps with a learning rate of 2e-5 and a batch size of 256.
We use EMA of 0.9999.
We keep the CLIP encoder for the textual prompt and the VAE frozen during training.

\section{Results}
\label{sec:result_and_analysis}

\subsection{Model Bias}

\begin{figure*}[ht]
\centering
\includegraphics[width=1\linewidth]{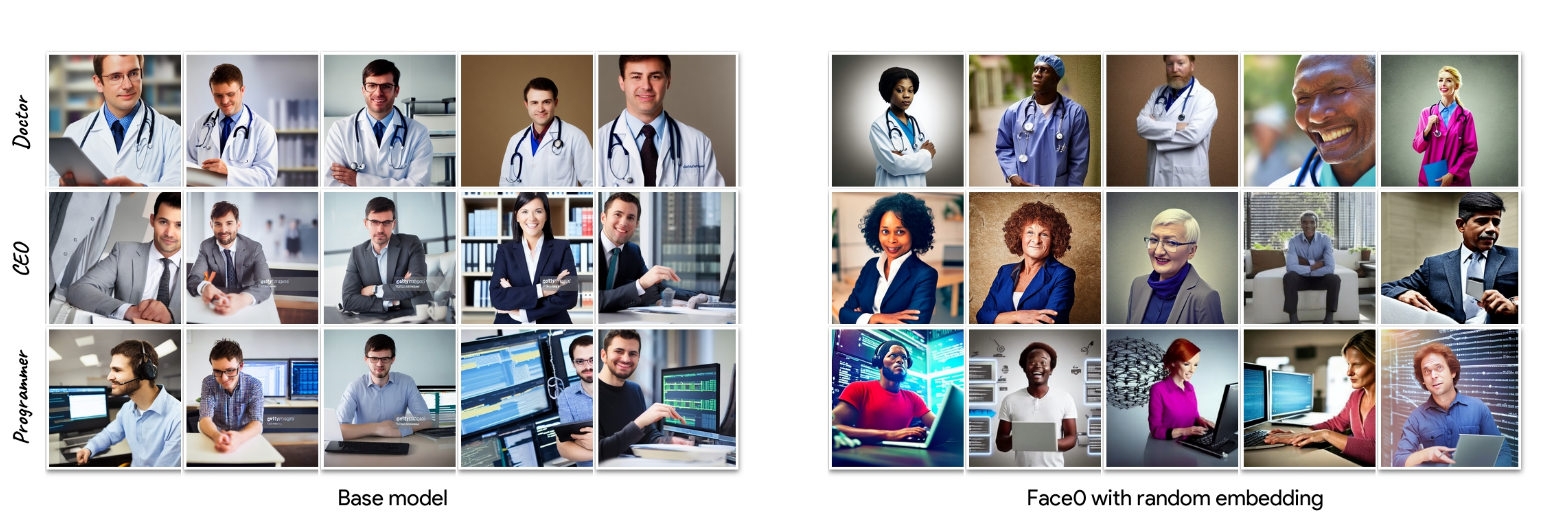}
\caption{Samples for the prompts ``A stock photo of X'' for X in $\{``doctor", ``CEO", ``programmer"\}$ from the base model (left) and our model with a random face embedding (right).}
\label{fig:bias}
\end{figure*}

Text-to-image diffusion models may inherit unfair biases from their underlying training data. One type of bias might be correlating facial features with specific words unrelated to facial features. Since our model is incentivized to decouple facial features from textual prompts, a simple variation of our method might allow some mitigation of this type of bias. Specifically, instead of taking a face embedding vector generated by our face embedding module from a given image, we can instead, for every generation from the model, use a randomly sampled face embedding vector. This procedure conditions the model on the random embedding which is decoupled from the biases the model might have for the textual prompts (see \cref{fig:bias}). We note that this doesn't affect the model's running time, and can be applied horizontally to all generations containing a face. This is only a preliminary result, and there are many important questions still open, for example what biases might exist within the face embeddings themselves, how the biases between the face embeddings and the textual prompt interact, and how to sample the random face embedding (for our experiments we used a simple mixture of Gaussians). In spite of these shortcomings, we hope that our preliminary results encourage further research on this important topic.

\subsection{Consistent Characters}

When generating images with a diffusion model, a user might encounter a character that they like. Unfortunately it is usually non-trivial to recreate the same character with the base model. With Face0, consistent character creation is trivial - we can just randomize an embedding when generating the character, and maintain one that we like for maximum fidelity, or simply calculate the face embedding from an image generated without one.

We can also do the reverse: if we get a generation that we like but would like to use a different face, we can keep the latent seed and prompt fixed, and only change the conditioning embedding. While this sometimes results in larger changes, often the main effect would be to just change the face (see \cref{fig:fig1}).

\subsection{Controllability}
 
\begin{figure*}[ht]
\centering
\includegraphics[width=.8\linewidth]{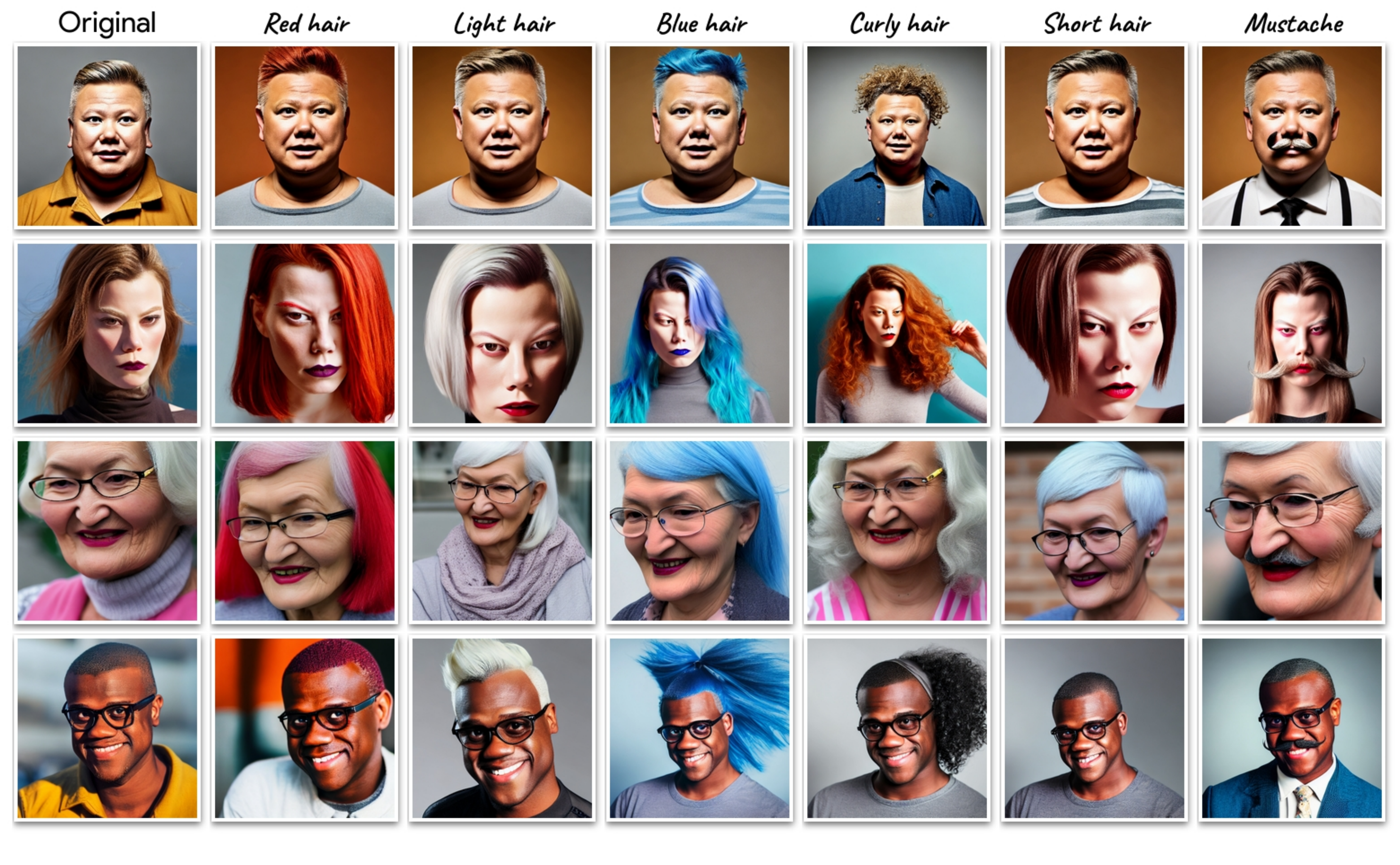}
\caption{Our model allows overriding features from the face embedding via the textual prompt.}
\label{fig:prompt_control}
\end{figure*}

\begin{figure*}[ht]
\centering
\includegraphics[width=.8\linewidth]{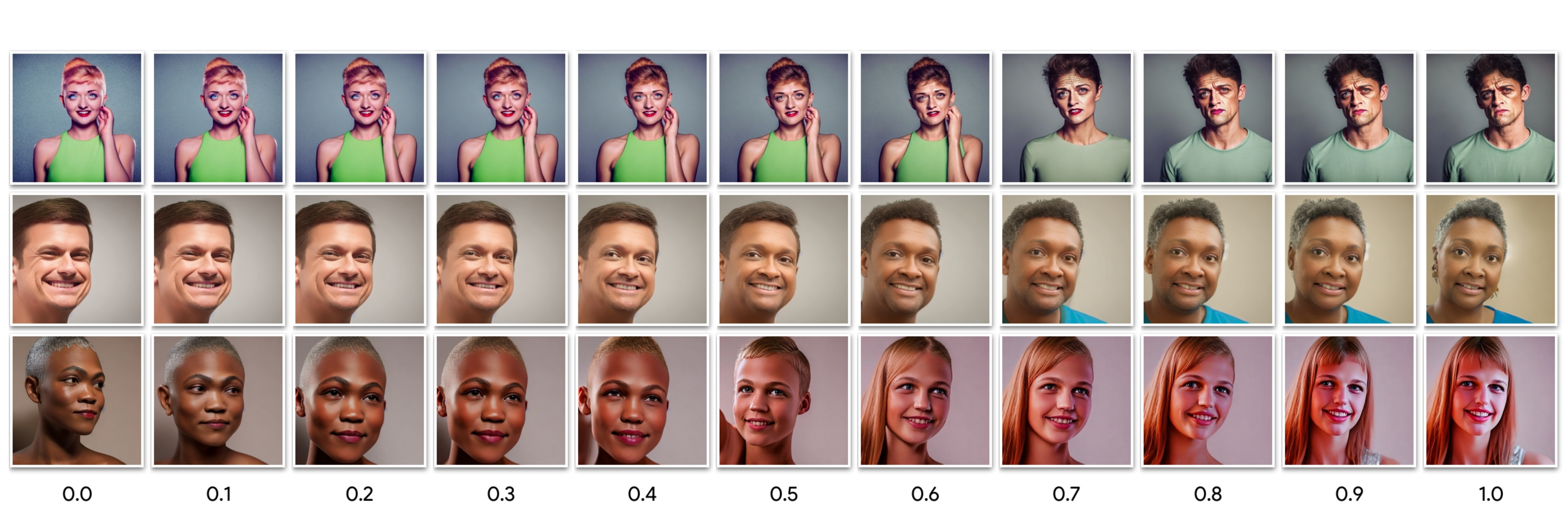}
\caption{Face0 enables control of facial features that are harder to describe textually via direct manipulation of the face embedding. Here we see simple linear interpolation between the left and right faces.}
\label{fig:lerp}
\end{figure*}

The embedding mechanism we use in Face0 mostly fixes facial features, pose and expression when no other conditions are provided. However, our model allows controlling the generated face in two ways. First, we can modify the generated faces by using the textual prompt and specifying a trait that contradicts the embedding. For example, ``a person with blue hair'' (see \cref{fig:prompt_control}). Second, our model provides additional controllability for features that are harder to describe textually. For example, with a simple linear interpolation between two face embeddings, we can create a meaningful semantic transition between the faces (see \cref{fig:lerp}). Note that interpolation also provides a simple way to take into account several face images of the same person. For example, we can average the embedding or do a weighted average with different weights between several images of the same person. Since generation is immediate, the weights can be tuned interactively. We only did minimal experimentation with this but this showed promising results for further research.

\begin{figure*}[ht]
\centering
\includegraphics[width=.75\linewidth]{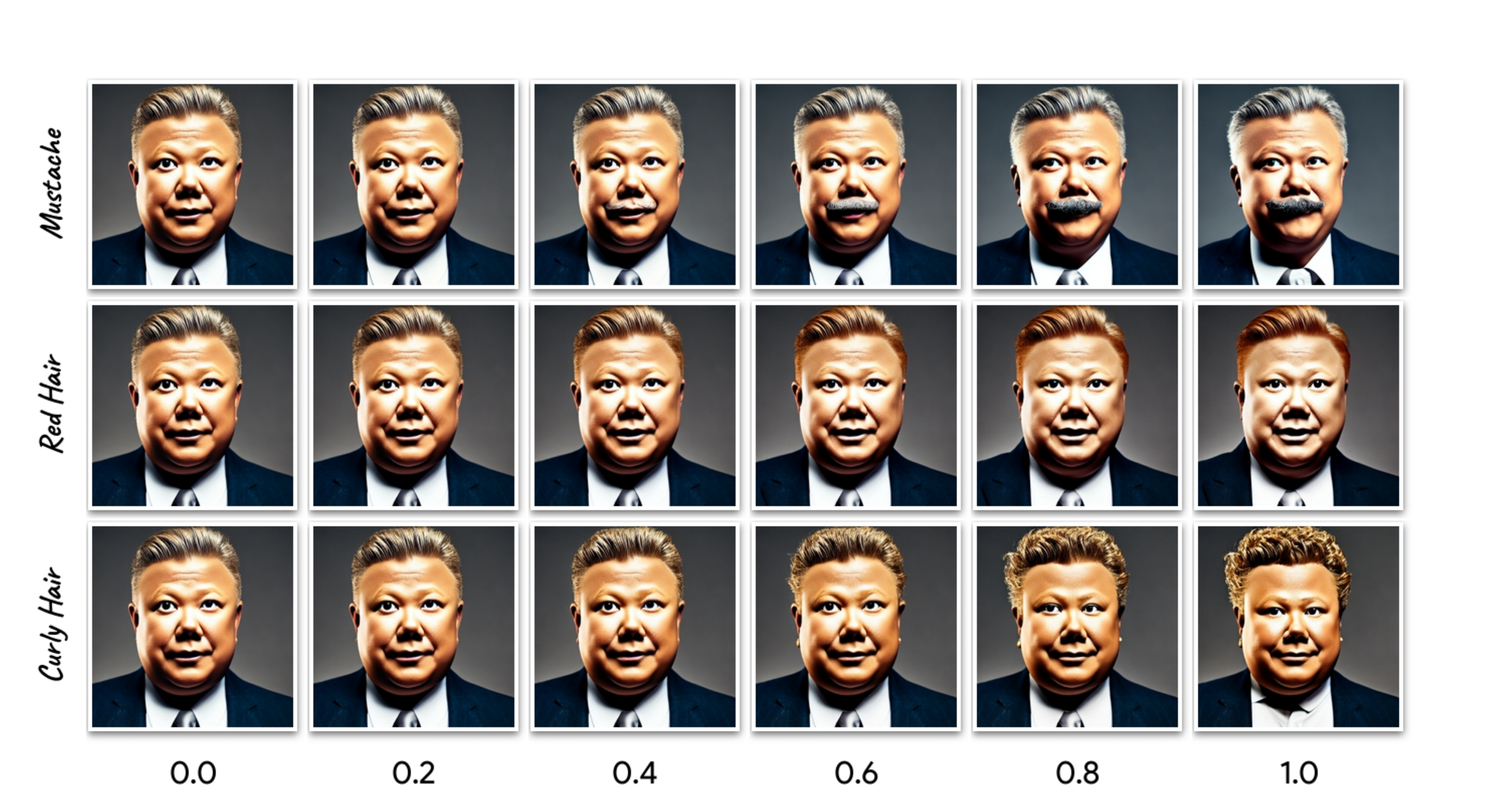}
\caption{Face0 enables fine-grained control of facial features that are harder to describe textually via direct manipulation of the face embedding. Here we see a simple linear interpolation between the facial embeddings of two generated photos from the same source (the top-left image in \cref{fig:prompt_control}) with different textual prompts.}
\label{fig:text_and_face_lerp}
\end{figure*}

Finally, we can combine the two control methods to attenuate the strength of the textual control. For example, we can create an image with a face embedding $e_{src}$ and the prompt ``a person with a mustache,'' then calculate the embedding $e_{mustache}$ and, using linear interpolation between the embeddings, control the amount of mustache (See \cref{fig:text_and_face_lerp}).

\subsection{Sampling Variations}
\label{sec:sampling}

When generating images with Face0, we found that changing the weights of the facial, textual and combined embeddings independently allowed for fine-grained control over the resulting images in useful ways. As mentioned above in our experiments we fixed the CFG weight as $w=7.5$. When a photo-realistic result was desired (ex. superheroes, doctors in \cref{fig:fig1}) we set $c=1$, essentially doing standard CFG; in other words, all of the CFG strength was given to the combined vector. When non-photo-realistic image generations were desired (ex. the Van Gogh style paintings or the action figures in \cref{fig:fig1}) we simply increased the weight of the textual embedding over the combined embedding. Values in the range $0.4 < c < 0.7$ and $a = 0$ work well. We also experimented with assigning a negative weight to the facial embedding, and increasing the weight of the combined vector, e.g. $c=1.4, a=1$, for very non-photo-realistic images.

\begin{figure}[ht]
\centering
\includegraphics[width=1\linewidth]{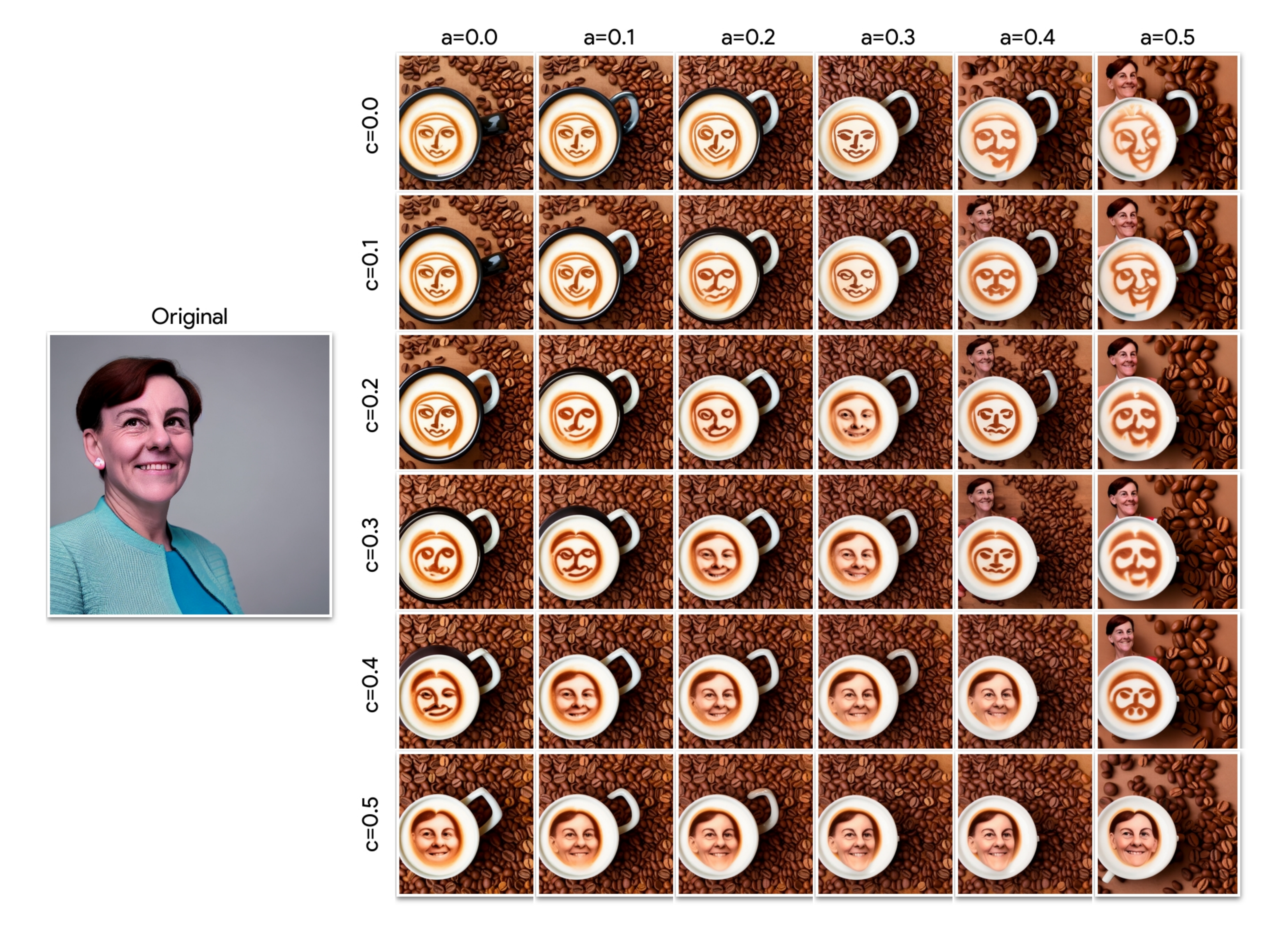}
\caption{Face0 can independently weight the textual, facial, combined and unconditioned embeddings. All images were generated with the prompt "latte art of a face in a mug" and a fixed latent seed. See \cref{sec:sampling} for details.}
\label{fig:cfg_weights_variance}
\end{figure}

\cref{fig:cfg_weights_variance} illustrates the effect of changing the weights. As $c$ increases, we give more weight to the combined embedding and as $a$ increases, we shift weight from the textual embedding to the facial embedding.

\section{Comparisons}
\label{sec:comparisons}
We compare our method to Dreambooth \cite{Dreambooth}, a method that conditions a diffusion model on a given subject by fine-tuning it on 3-5 images of that subject. One key difference between the methods is inference time -- training a Dreambooth model on a subject took 15 minutes on an A100 GPU, while Face0 does not require per-subject training.

\subsection{Dataset}
To perform the comparison, we created a dataset of 20 synthetic identities, with multiple photos for each identity (we extracted faces from these photos as described in \cref{sec:method}). In addition, we used 10 identities from the Labeled Faces in the Wild (LFW) \cite{LFW} dataset. We tested the models on 10 prompts (5 photo-realistic and 5 artistic). Dreambooth was trained on all available images (4-5) and Face0 received only a single photo as input. We collected results from both methods on 8 random seeds using standard DDIM sampling (i.e. we used $c=1$ for Face0).

\subsection{Qualitative results}

\begin{figure}[ht]
\centering
\includegraphics[width=1\linewidth]{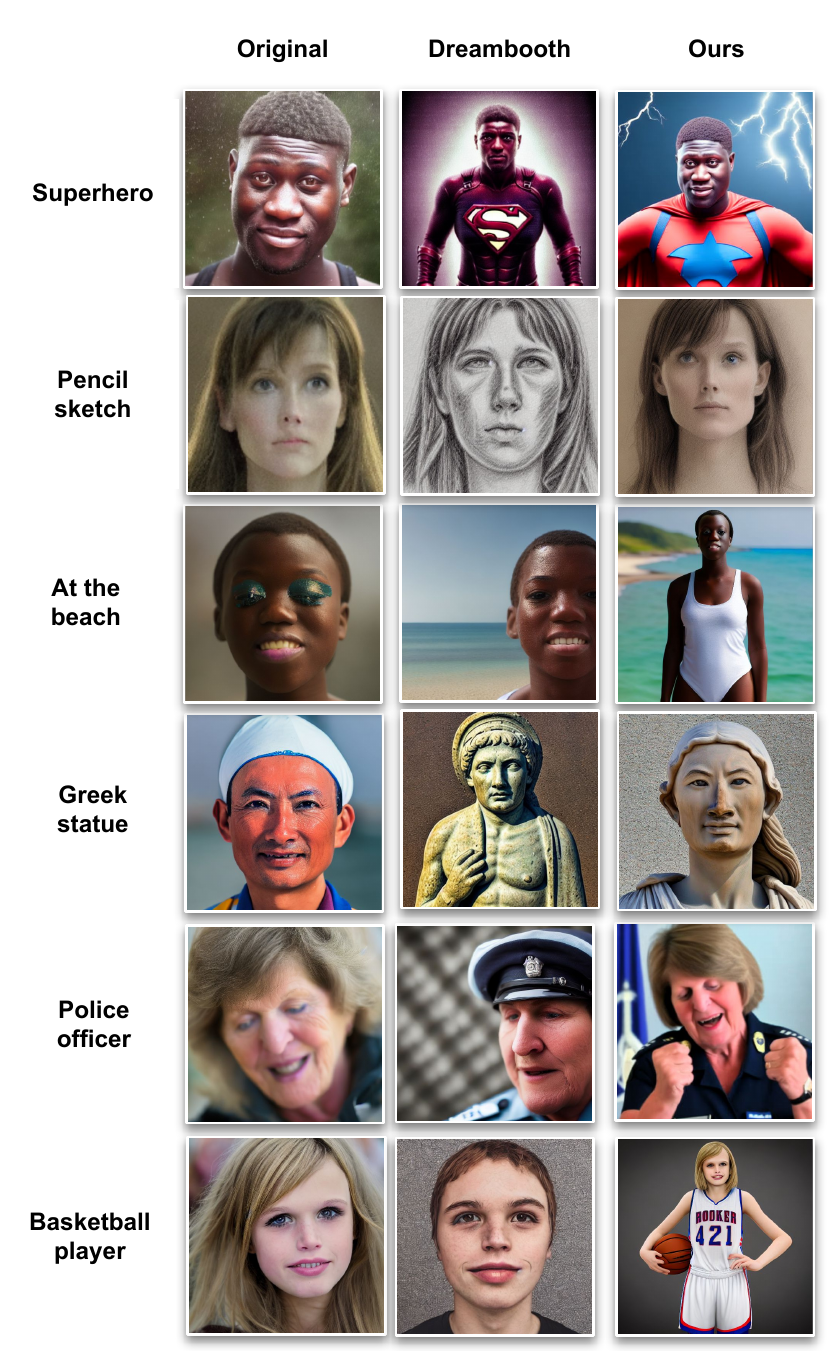}
\caption{Comparison of Face0 and Dreambooth. Images shown are the best out of 8 random samples.}
\label{fig:comparison}
\end{figure}

Qualitative results can be seen in \cref{fig:comparison}. We observe comparable results, and note that it was easier to get consistent appearance of the face when using Face0 (Dreambooth often ignored the face altogether requiring multiple seeds to obtain a good result). On the other hand, when there is an identity mismatch in Face0, it is preserved across seeds.

\subsection{Quantitative results}
To perform a quantitative evaluation of the methods we measure alignment with the provided text and the provided face. For text alignment, we measure the cosine similarity between the CLIP \cite{CLIP} embeddings of the generated image and the textual prompt. For face alignment, we extract the largest face from the generated image and compare it with a face provided to the model using cosine similarity of the CLIP embeddings. When no faces exist in the generated image, we set the similarity to 0. We define the overall score of each generated image as the sum of the face and text scores.

Both methods perform comparably on text alignment, but Face0 scored better at aligning with the provided face (\cref{table:comparison1}). In addition, both methods performed slightly better on synthetic images. To verify our qualitative observations we also measured the performance of the methods when considering the best generation out of the 8 random seeds (\cref{table:comparison2}). We see a performance improvement for Dreambooth, showing that it's less consistent than Face0.

\begin{table}
\centering
\caption{Quantitative comparison of Face0 and Dreambooth over synthetic faces (SYN) and faces from the LFW dataset (LFW).}

{
    \begin{tabular}{|c c c c| } 
     \hline
     Method & Text Align.  & Face Align.  & Overall \\
     \hline

     Face0 (SYN)  & $ 0.24 \pm 0.02 $ & $ 0.72 \pm 0.07 $ & $ 0.96 \pm 0.07 $  \\ 
     DreamBooth (SYN)  & $ 0.23 \pm 0.03 $ & $ 0.46 \pm 0.19 $ & $ 0.69 \pm 0.18 $ \\

     \hline

     Face0 (LFW)  & $ 0.23 \pm 0.03 $ & $ 0.66 \pm 0.06 $ & $ 0.89 \pm 0.06 $  \\ 
     DreamBooth (LFW)  & $ 0.24 \pm 0.02 $ & $ 0.39 \pm 0.14 $ & $ 0.62 \pm 0.13 $ \\

     \hline
    \end{tabular}

}
\label{table:comparison1}
\end{table}

\begin{table}
\centering
\caption{Overall scores for Face0 and Dreambooth when selecting the best score out of 8 seeds vs. the average.}
\begin{tabular}{|c c c|} 
 \hline
 & Face0 & DreamBooth \\
 \hline
 Best (SYN) & $ 1.04 \pm 0.06 $ & $ 0.93 \pm 0.11 $ \\
 Average (SYN) & $ 0.96 \pm 0.07 $ & $ 0.69 \pm 0.18 $ \\
 \hline
 Best (LFW) & $ 0.98 \pm 0.04 $ & $ 0.83 \pm 0.10 $ \\
 Average (LFW) & $ 0.89 \pm 0.06 $ & $ 0.62 \pm 0.13 $ \\

 \hline
\end{tabular}
\label{table:comparison2}
\end{table}

\section{Related work}
\textbf{Text-to-Image Models.} Deep generative models for image generation have shown tremendous progress in recent years. Early approaches relied on training a GAN \cite{GAN} generator (like StyleGAN \cite{StyleGAN}) and guiding it using CLIP \cite{CLIP} in various methods \cite{Clip2StyleGAN, StyleCLIP}. More recently, transformer-based methods \cite{DALLE1, Parti,muse} and diffusion models \cite{DallE2,StableDiffusion,Imagen} have gained popularity as they allow easy text-conditioning and can generalize to broader domains. Our work, demonstrated on the Stable Diffusion model \cite{StableDiffusion}, shows that diffusion models can also be easily conditioned on other modalities, such as face encoding.

\textbf{Image embedding.} Encoding image pixels into a latent representation that contains useful features for downstream models is an important and long-standing problem in deep learning. Some useful encoding are achieved by training an autoencoder \cite{VQVAE,VQGAN}, while other methods take an intermediate layer of a model that was trained to solve an image-related problem like image recognition \cite{VIT}. A popular image encoder is CLIP \citep{CLIP} which is obtained by aligning images with textual captions. In our work we use an intermediate layer of an Inception Resnet \cite{inception-v1,inception-v4} that was trained on the vggface2 \citep{vggface2} dataset.

\textbf{Personalization of image generation models.} Personalized image generation attempts to include new subjects, described by one or more images, in the resulting synthesized image. A common approach to this problem is to fine-tune an image generation model during inference, on the provided images. MyStyle \cite{MyStyle} fine tunes a StyleGAN \citep{StyleGAN} on a custom face image. DreamBooth \cite{Dreambooth} and Textual Inversion \cite{TextualInversion} enable personalization in diffusion models using fine-tuning (either of the model itself or of an entry in the embedding table of the textual encoder). Other methods \cite{unitune,dreamix} use fine-tuning for text-guided editing of a single input image or video. These methods perform fine-tuning during inference, and are therefore costly in time and memory. More recent advancement \cite{LORA-dreambooth} use LORA \cite{LORA} to address the memory cost, but speed is still an issue. Concurrently with our work, \cite{RinonEncoder} suggest to use intermediate layers in CLIP \citep{CLIP} as input to an image encoder. This significantly lowers the amount of fine-tuning required for high quality inference, but does not eliminate it.

\section{Discussion and Limitations}
\label{sec:discussion}

We presented Face0, a novel and simple method for conditioning a diffusion based image generation model on a face.
Once trained, the model is able to produce pleasing results extremely quickly, practically at the same cost as the base model.
Our method allows for controlling more or less photo-realistic generations (by varying the text-only, face-only and combination CFG weights balance).
We show that it is easy to override properties of the face embedding with the textual prompt.
We also show that our method can help solve the problem of consistent character generation, by keeping a fixed face embedding vector.
Finally, while more research is needed, we show that training the model to decouple its textual conditioning from its conditioning on a face, is hopefully a step towards some mitigation of some of the biases of the base model.

There are several interesting related directions for further research and improvements. For example, we choose a face embedding mechanism that mostly fixes the face pose and expression, but it would be interesting to experiment with other face embedding mechanisms. In addition, while generating pleasing results, Face0 is not always able to fully preserve a provided identity, and sometimes creates "look-alike" characters that are close in appearance but still distinguishable from the input face. We are hopeful that this can be improved by smart noising of the embedding vector and by experimenting with conditioning the model on multiple faces (in our experiments we only allowed one) which we leave for future work. Another interesting idea would be to use the face embedding model to guide sampling at each sampling step. Finally, while faces are especially important to condition on, it might be interesting to apply the same method to additional domains.

\section*{Societal impact}

Face0, like other image generation techniques, has a great potential to complement and augment human creativity by creating new tools for professionals and empowering non-professionals with the ability to create images more easily and in a more intuitive manner. However, we recognize that applications of this research may impact individuals and society in complex ways (see \cite{Imagen} for an overview). In particular, this method illustrates the ease with which such models can be used to alter sensitive characteristics such as skin color, age and gender. Although this has long been possible by means of image editing software, text-to-image models can make it easier.

Another cause of concern is reproducing unfair bias that may be found in the underlying model training data. This is also relevant for the underlying model, Stable Diffusion. Moreover, these unfair biases may make the performance of the model vary across people of different groups. While we did not see this effect in our qualitative experiments, more research into bias evaluation methods, both for image editing and generation, will help address this concern. In addition, while these capabilities already exist in image editing software, for example, single image personalization methods, such as Face0, may increase the ability to forge convincing images of non-public individuals, or make it easier to generate disinformation and manipulate images in hateful and harassing ways.

We encourage future research to help mitigate and measure the potential negative impact of generative models if misused, and believe thoughtful consideration and further research in all of these matters is necessary prior to determining how such technologies can be made broadly available.

\section*{Acknowledgments}
We would like to thank Matan Kalman, Jason Baldridge, Kathy Meier-Hellstern, Tom Duerig, Caroline Pantofaru, Michael Nechyba, Dmitry Lagun, Viral Carpenter, Eyal Segalis, Eyal Molad, Yael Pritch, Shlomi Fruchter, the Theta Labs team at Google, and our families.

\bibliographystyle{ACM-Reference-Format}
\bibliography{references}

\begin{figure*}[ht]
\centering
\includegraphics[width=1\linewidth]{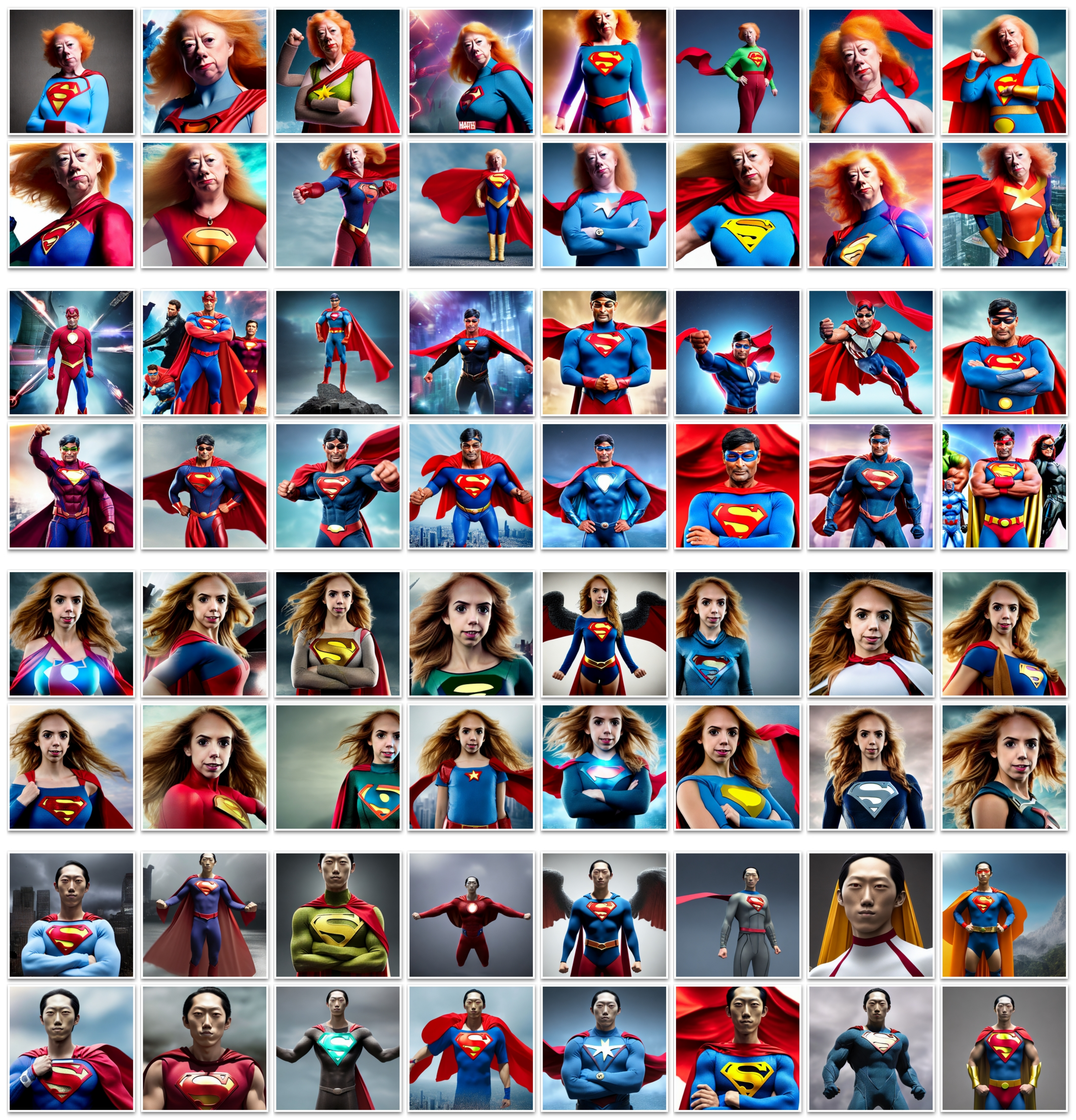}
\caption{Face 0 maintains consistency across generations. Non-cherry picked examples using the face embeddings from the original photos in \cref{fig:fig1}.}
\label{fig:consistency}
\end{figure*}

\begin{figure*}[ht]
\centering
\includegraphics[width=1\linewidth]{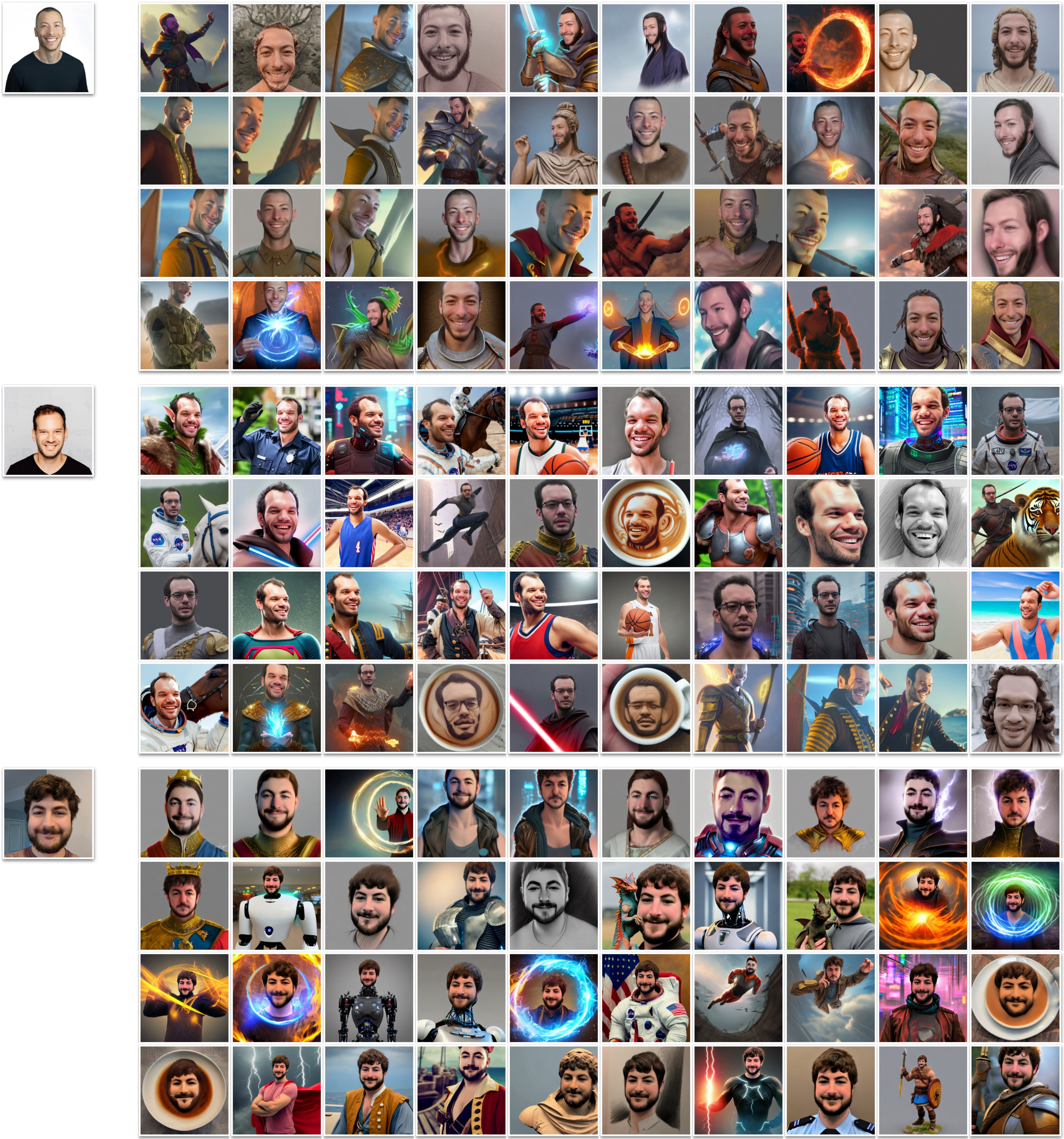}
\caption{Additional examples using several images of the same person and a variety of prompts.}
\label{fig:consistency}
\end{figure*}

\end{document}